\documentclass[10pt,twocolumn,letterpaper]{article}

\usepackage{iccv}
\usepackage{times}
\usepackage{epsfig}
\usepackage{graphicx}
\usepackage{amsmath}
\usepackage{amssymb}
\usepackage{capt-of}

\usepackage[utf8]{inputenc} 
\usepackage[T1]{fontenc}    
\usepackage{booktabs}       
\usepackage{amsfonts}       
\usepackage{nicefrac}       
\usepackage{microtype}      
\usepackage{xcolor}         
\usepackage{bm}

\usepackage{bbding}
\usepackage{tabularx}
\usepackage{booktabs}
\usepackage{multirow}
\usepackage{xspace}
\usepackage[acronym]{glossaries}
\usepackage{float}
\usepackage{wrapfig}

\definecolor{blueish}{rgb}{0.0, 0.3, .6}

\newcommand{\AT}[1]{{\color{blueish}{\bf [AT: #1]}}}

\renewcommand{\AT}[1]{}

\usepackage{amsmath,amsfonts,bm}

\renewcommand{\paragraph}[1]{\vspace{.5em}\noindent\textbf{#1}.}

\def\eqref#1{equation~\ref{#1}}

\def\1{\bm{1}}

\DeclareMathAlphabet{\mathsfit}{\encodingdefault}{\sfdefault}{m}{sl}
\SetMathAlphabet{\mathsfit}{bold}{\encodingdefault}{\sfdefault}{bx}{n}

\newcommand{\kitti}{KITTI-360\xspace}
\newcommand{\threedfront}{3D-FRONT\xspace}

\usepackage[pagebackref=true,breaklinks=true,letterpaper=true,colorlinks,bookmarks=false]{hyperref}

\iccvfinalcopy 

\def\iccvPaperID{11606} 

\begin{document}

\title{CC3D: Layout-Conditioned Generation of Compositional 3D Scenes}

\author{
  Sherwin Bahmani$^{*1}$
  \space\space
  Jeong Joon Park$^{*2}$ 
  \space\space
  Despoina Paschalidou$^{2}$ 
  \space\space
  Xingguang Yan$^{4}$\\
  Gordon Wetzstein$^{2}$ 
  \space\space
  Leonidas Guibas$^{2}$ 
  \space\space
  Andrea Tagliasacchi$^{1,3,4}$\\
  \\
  \space\space
  \textnormal{$^{1}$University of Toronto \space\space$^{2}$Stanford University\space\space$^{3}$Google Research \space\space$^{4}$Simon Fraser University}
}

\twocolumn[{%
\renewcommand\twocolumn[1][]{#1}%
\maketitle


\begin{center}
    \centering
    \vspace{-0.6em}
    \includegraphics[width=0.9\linewidth]{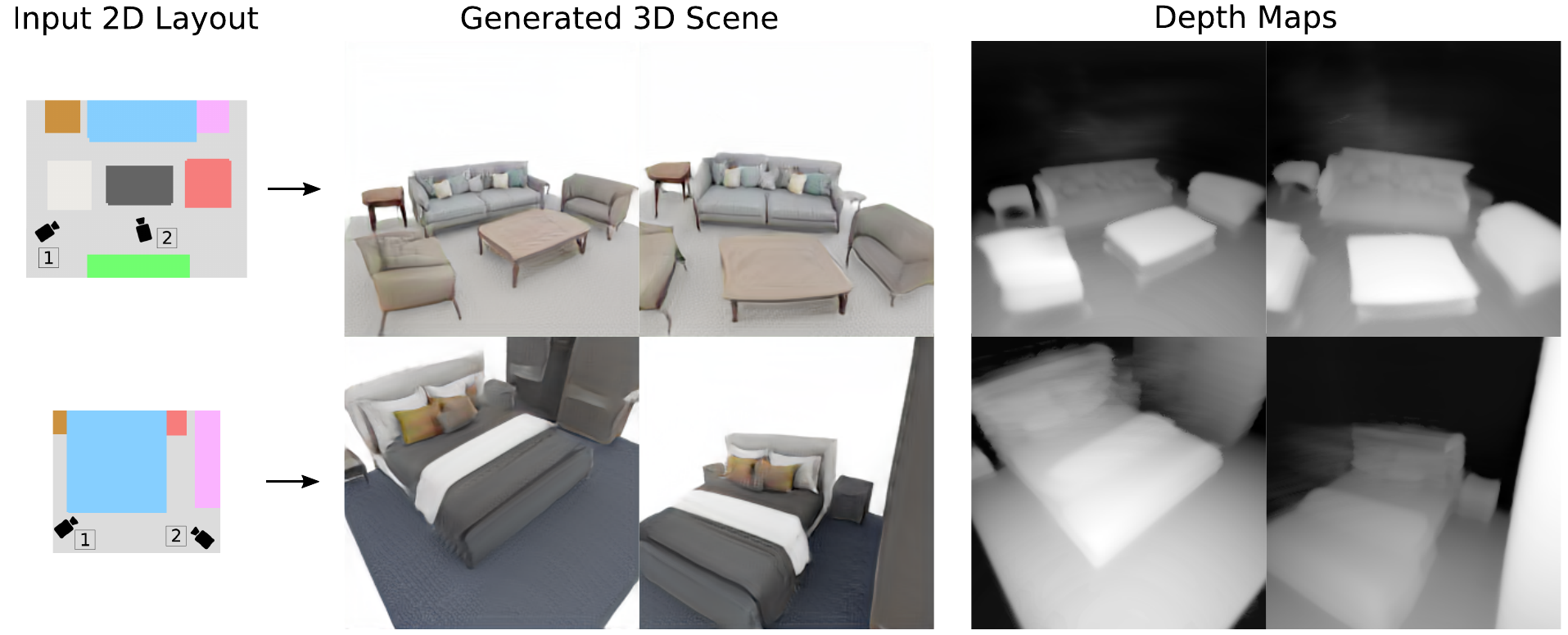}
    \captionof{figure}{\textbf{Compositional 3D Scene Generation} -- We introduce CC3D, a 3D compositional GAN capable of synthesizing view-consistent renderings of multi-object scenes conditioned on a semantic layout, describing the scene structure. Video results can be viewed on our website: \url{https://sherwinbahmani.github.io/cc3d}. 
    \vspace{-0.2em}
    }
    \label{fig:teaser}
\end{center}%
}]

\begin{abstract}
\vspace{-0.8em}
In this work, we introduce CC3D, a conditional generative model that synthesizes complex 3D scenes conditioned on 2D semantic scene layouts, trained using single-view images. Different from most existing 3D GANs that limit their applicability to aligned single objects, we focus on generating complex scenes with multiple objects, by modeling the compositional nature of 3D scenes. By devising a 2D layout-based approach for 3D synthesis and implementing a new 3D field representation with a stronger geometric inductive bias, we have created a 3D GAN that is both efficient and of high quality, while allowing for a more controllable generation process. Our evaluations on synthetic \threedfront and real-world \kitti datasets demonstrate that our model generates scenes of improved visual and geometric quality in comparison to previous works.
\def\thefootnote{*}\footnotetext{Equal contribution}\def\thefootnote{\arabic{footnote}}
\vspace{-0.5em}

\end{abstract}
\vspace{-1em}

\section{Introduction}
\label{sec:intro}

\begin{figure}[t]
\begin{center}
\centerline{\includegraphics[width=1.0\linewidth]{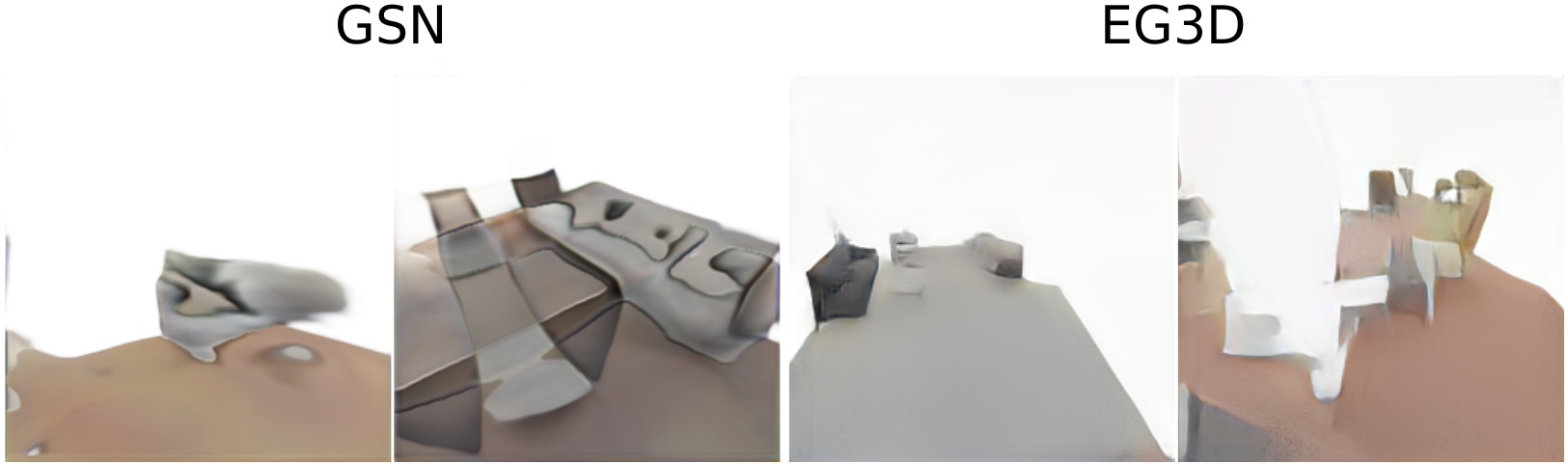}}
\caption{\textbf{Failure cases}  -- of non-compositional 3D GANs on multi-object scenes. We show examples of generated scenes synthesized with \cite{devries2021unconstrained, chan2022efficient}, which do not explicitly model the compositional nature of multi-object living rooms.  Note their lack of visual qualities.
}
\label{fig:low_quality}
\end{center}
\vskip -0.4in
\end{figure}

Recently, we have witnessed impressive progress in 3D generative technologies, including generative adversarial networks (GANs)~\cite{goodfellow2014generative} that have emerged as a powerful tool for automatically creating realistic 3D content.
 Despite their impressive capabilities, existing 3D GAN-based approaches have two major limitations. First, they typically generate the entire scene
 from a single latent code, ignoring the compositional nature of multi-object scenes, thus struggling to synthesize scenes with multiple objects, as shown in Fig.~\ref{fig:low_quality}.
 Second, their generation process remains largely uncontrollable, making it non-trivial to enable user control.
 While some works \cite{cai2022pix2nerf, lin20223d} allow conditioning the generation of input images via GAN inversion, this optimization process can be time-consuming and prone to local minima.

In this work, we introduce a {\bf{C}}ompositional and {\bf{C}}onditional {\bf{3D}} generative model (\textbf{CC3D}), that generates plausible 3D-consistent scenes with multiple objects,
while also enabling more control over the scene generation process by conditioning on semantic instance layout images, indicating the scene structure (see Fig.~\ref{fig:teaser}). Our approach rhymes with the 2D image-to-image translation works \cite{isola2017image,epstein2022blobgan} that conditionally generate images from user inputs: CC3D generates 3D scenes from 2D user inputs~(\ie scene layouts).

To train CC3D we use a set of single-view images and top-down semantic layout images, such as 2D labelled bounding boxes of objects in a scene (\eg Fig.~\ref{fig:teaser}). Our generator network takes a 2D semantic image as input that defines the scene layout and outputs a 3D scene, whose top-down view matches the input layout in terms of object locations.

The key component of our approach is a 2D-to-3D translation scheme that efficiently converts the 2D layout image into a 3D neural field. Our generator network is based on a modified StyleGAN2~\cite{karras2020analyzing} architecture that processes the input 2D layout image into a 2D feature map.
The output 2D feature map is then reshaped into a 3D feature volume that defines a  neural field which can be rendered from arbitrary camera views.
Similar to existing 3D-aware generative models \cite{schwarz2020graf, niemeyer2021giraffe, chan2022efficient}, we train the generator to produce realistic renderings of the neural fields from all sampled viewpoints. In addition, we enforce a semantic consistency loss that ensures the top-down view of the 3D scene matches the semantic 2D layout input.

\begin{figure*}
\centering
\includegraphics[width=1.0\linewidth]{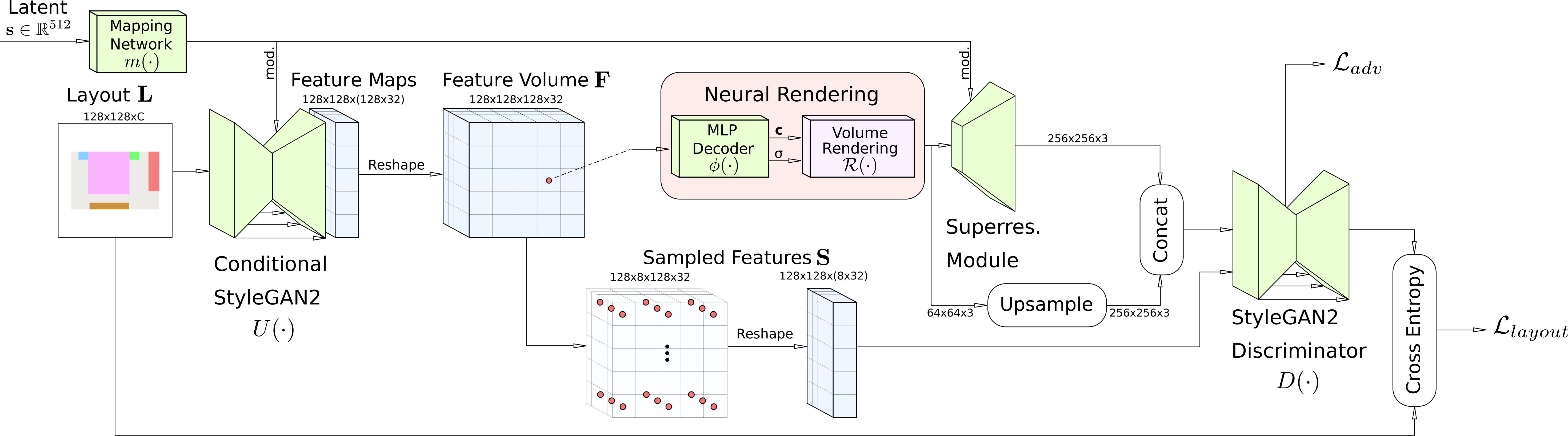}
\caption{\textbf{Architecture} --
Our method takes a floorplan projection of the semantic scene layout and a noise vector as inputs. We use a conditional StyleGAN2 backbone to generate a 2D feature field based on the given layout and reshape the channels into a 3D feature volume. This feature volume is queried using trilinear interpolation and subsequently decoded into color and density using a small MLP. We use a superresolution module to upsample volume rendered images to target resolution and use a standard StyleGAN2 discriminator. In order to ensure semantic consistency between the layout and the rendering, we sample equidistant coordinates from the feature volume and process the sampled features with a semantic segmentation decoder added to the discriminator. We train our model on a combination of an adversarial loss and cross entropy loss.
}
\label{fig:method}
\vspace{-0.5cm}
\end{figure*}

We evaluate CC3D on the 3D-FRONT~\cite{fu20213d} bedroom and living room scenes and the KITTI-360 dataset~\cite{liao2022kitti} that contains more challenging outdoor real-world scenes.
Our evaluations demonstrate that existing 3D generative models, such as EG3D~\cite{chan2022efficient} and GSN~\cite{devries2021unconstrained}, produce low-quality 3D scenes, as illustrated in Fig.~\ref{fig:low_quality}.
In comparison, the compositional generation process and the new intermediate 3D feature representation of CC3D significantly improve the fidelity of the synthesized 3D scenes on both datasets, opening the door for realistic multi-object scene generations.

\section{Related Work}
\paragraph{2D Image Synthesis}
GANs~\cite{goodfellow2014generative} have been extensively utilized to generate photorealistic images \cite{karras2019style, karras2020analyzing, karras2021alias, brock2018large, sauer2022stylegan}, perform image-to-image translation \cite{isola2017image, zhu2017unpaired, choi2018stargan}, and image editing \cite{wang2018high, shen2020interpreting, ling2021editgan}. Recently, compositional approaches \cite{hudson2021generative, arad2021compositional} have also been explored in the context of image generation. Similar to our work, GANformer2~\cite{arad2021compositional} also divides the generation process into two steps: planning and execution. In our work, we guide the {\em 3D} generation process using semantic layouts and demonstrate
that CC3D can render multi-view consistent images of multi-object scenes.

\paragraph{3D Object Generation}
To scale 2D GANs to 3D domain,
many recent works explored combining image generators with 3D representations.
These models are supervised only with unstructured image collections along with a pre-defined camera distribution. While earlier works \cite{nguyen2019hologan, nguyen2020blockgan, schwarz2020graf, chan2021pi, niemeyer2021campari} provided limited visual fidelity and geometric accuracy, recently, several works tried to
address these limitations. The majority of these approaches \cite{zhou2021cips, deng2022gram, gu2022stylenerf, or2022stylesdf, chan2022efficient, xu20223d, skorokhodov2022epigraf,gao2022get3d} use a style-based generator \cite{karras2020analyzing} to synthesize a neural field which can be used for volume rendering \cite{max1995optical}. Although these approaches can produce high quality images for single-object scenes, they fail to scale to complex scenes with multiple objects. In this work, we also employ a style-based generator in combination with volume rendering but as our model explicitly models the compositional nature of 3D scenes, it can successfully generate plausible indoor and outdoor 3D scenes.

\paragraph{Multi-Object Generation}
Our work is closely related to recent approaches that model scenes using 3D-aware image generators
\cite{niemeyer2021giraffe, xue2022giraffe}. Among the first, GIRAFFE \cite{niemeyer2021giraffe} proposed to represent scenes using multiple locally defined NeRFs. However, while \cite{niemeyer2021giraffe} can be efficiently applied on scenes containing only a few objects with limited texture variation, such as the CLEVR \cite{johnson2017clevr} dataset, it fails to generalize to more complex scenes. To improve the
visual quality of \cite{niemeyer2021giraffe},
GIRAFFE-HD \cite{xue2022giraffe} employed a style-based generator. Even though this allows their model to composit multiple objects of the same class, e.g., cars, into a single scene at inference time, learning compositional scene generation from multi-object scenes of different classes remains an open problem.

\paragraph{Large-Scale Scene Generation}
Plan2Scene \cite{vidanapathirana2021plan2scene} focuses on the task of converting a floorplan accompanied by a sparse set of images into a textured mesh for the entire scene.
Although their representation is compositional by construction, \cite{vidanapathirana2021plan2scene} is not generative and requires  multi-view supervision.
Closely related to our work, another line of research \cite{devries2021unconstrained, bautista2022gaudi} aims at generating large-scale scenes using locally conditioned neural fields. Unlike previous approaches that sample camera poses from a sphere targeted towards the origin, constraining them to $S O(3)$, GSN \cite{devries2021unconstrained} considers scene generation conditioned on a freely moving camera defined in $S E(3)$.
Although this setup permits generating scenes from arbitrary viewpoints, it makes training significantly harder, as datasets are not aligned and the range of possible camera poses drastically increases. 
GAUDI \cite{bautista2022gaudi} further improves the quality by disentangling camera poses from geometry and appearance.
Unlike GAUDI \cite{bautista2022gaudi} that assumes multi-view input images with known camera poses, our model can be trained using unstructured set of images.

\paragraph{Indoor Scene Generation}
Recently, several works \cite{wang2018deep, ritchie2019fast, wang2021sceneformer, paschalidou2021atiss,wei2023lego} proposed to pose the scene generation task
as an object layout prediction problem. For example, ATISS \cite{paschalidou2021atiss} uses an autoregressive transformer to generate synthetic indoor environments as an unordered set of objects. LEGO-Net \cite{wei2023lego} learns to iteratively refine random object placements to generate realistic furniture arrangements.  These works represent a scene layout as a set of 3D labeled bounding boxes, which can be replaced with textured meshes from a dataset of assets. In contrast, we rely on a GAN to learn a mapping between a 2D compositional scene layout to a 3D scene, without having to rely on object retrieval to produce 3D objects. We see our work as an orthogonal work to \cite{wang2018deep, ritchie2019fast, wang2021sceneformer, paschalidou2021atiss} as they can be used to generate scene layouts, which in turn can be used as our conditioning.

\paragraph{Concurrent Works}
Several concurrent works explored extending 3D GANs to more complex scenarios. 
3DGP \cite{skorokhodov20233d} tackles non-aligned datasets by incorporating depth estimation and a novel camera parameterization, but their model focuses only on single objects.
SceneDreamer \cite{chen2023scenedreamer} generates unbounded landscapes from 2D image collections and semantic labels. However, their model is supervised with a ground truth height field, whereas we learn the density field only from 2D image collections.
InfiniCity \cite{lin2023infinicity} synthesizes large-scale 3D city environments but requires expensive annotations such as CAD models.
Similar to ours, pix2pix3D~\cite{deng20233d} generates 3D objects given a 2D semantic map, but it only focuses on single-object scenes.
In concurrent work, DisCoScene \cite{xu2022discoscene} investigates compositional scene generation with layout priors using single-view image collections. Their approach follows \cite{niemeyer2021giraffe} and generates each object and the background independently. 
Unlike our work, DisCoScene conditions the scene generation on 3D layout priors, as opposed to 2D layouts, and assumes that the per-object attributes (\ie size, pose) are sampled from a pre-defined prior distribution. Instead, we do not assume this type of supervision. Moreover, unlike \cite{xu2022discoscene}, we explore rendering from freely moving cameras as opposed to cameras on a sphere. 
\section{Method}\label{sec:method}

The training of CC3D takes a set of single-view 2D RGB images and a set of top-down semantic layouts.
We \textit{do not} assume the two image sets to be in 1:1 correspondence;~Fig.~\ref{fig:method} illustrates the overall architecture of our method.

\paragraph{Training}
We randomly choose 2D semantic layout images and sample style codes.
The layout images and style codes are passed to the generator network, which outputs 2D features.
The 2D feature map is then reshaped into a 3D feature volume, which can be rendered via volume rendering with a small MLP network and a 2D upsampling network.
The realism of the rendered images is scored by the discriminator network against the set of RGB images, and the system is trained with the standard adversarial loss, along with the semantic consistency loss from the top-down views.

\paragraph{Inference}
We provide a semantic layout image and a style code to the generator to obtain a 3D neural radiance field that can be rendered from an arbitrary camera. Our method allows more control over the generation process compared to the most advanced unconditional GANs \cite{chan2022efficient, or2022stylesdf,gao2022get3d}, as users can specify the layouts with various styles and edit them.

\subsection{Neural Field Generator}
\label{sec:gen_main}
Our generator network $G(\textbf{L},\textbf{s})$ takes as input a 2D layout image $\textbf{L}$ and a style code $\textbf{s}\in \mathbb{R}^{512}$, sampled from a
unit Gaussian distribution, and generates a 3D neural feature field $\textbf{F}\in \mathbb{R}^{N\times N \times N\times C}$, where $N$ and $C$ correspond to the spatial resolution and channel size. In our experiments, we set $N=128$ and $C=32$.

\paragraph{Layout Conditioning}
\label{sec:conditioning}
The input to our generator is a 2D layout conditioning image $\textbf{L} \in \mathbb{R}^{N\times N\times L}$ that contains information about the scene structure, with $L$ being a dataset-dependent feature dimension.
In contrast to concurrent work \cite{xu2022discoscene} that uses 3D bounding boxes as conditioning, we choose to guide our generation with 2D semantic layouts images; this allows users to generate layouts via simple 2D editing instead.
The feature channels of an input layout are composed of the one-hot encoding of semantic classes and additional information such as bounding boxes' local coordinates, which is detailed in the supplementary. 
As illustrated in Fig.~\ref{fig:3dfront_2d}, conditioning the generation on a 2D semantic layout can allow us to conveniently control the structure and the style of a scene.

\begin{figure}[t]
\begin{center}
\centerline{\includegraphics[width=1.0\linewidth]{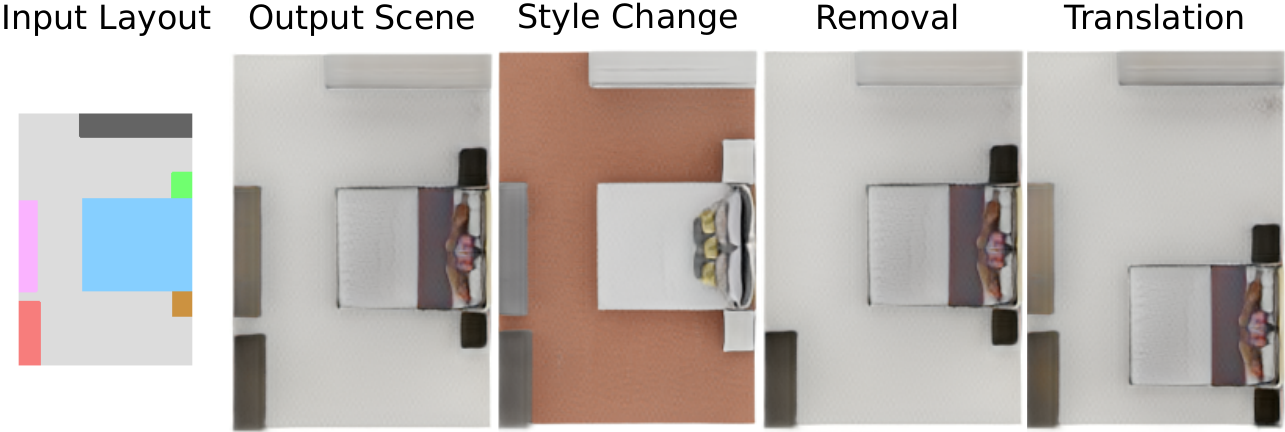}}
\caption{\textbf{Controllable scene generation} -- We conduct 2D experiments where we only train on top-down renderings. We observe that our model is able to follow a conditional layout while providing flexibility in varying the latent code or positioning of the objects.
}
\label{fig:3dfront_2d}
\end{center}
\vspace{-0.7cm}
\end{figure}

\paragraph{Layout-Conditioned 3D Generation}
\label{sec:generation}
As detailed in what follows, our conditional generator $G(\textbf{L},\textbf{s})$ is composed of a U-Net \textit{backbone} $U(\cdot)$ that generates a 2D feature image, followed by the \textit{extrusion} operator $E$ that reshapes a 2D feature map into a 3D feature grid as
\begin{equation}
    G(\textbf{L},\textbf{s})=E\circ U(\textbf{L},m(\textbf{s})).
\end{equation}

\paragraph{Backbone}
The network $U$ is a ``StyleGAN-like" U-Net architecture composed of encoder and decoder networks, and $m(\cdot)$ is a mapping network that conditions generation via FiLM~\cite{perez2018film}.
We use skip connections to concatenate the encoder features to the intermediate features of the corresponding decoding layer; please refer to the supplementary for additional details.
At the last layer, we have a single convolutional layer that increases the number of channels to a multiple of the height dimension of our target 3D feature volume.

\paragraph{Extrusion}
Finally, we convert the U-Net's 2D output in a 3D feature grid with the \textit{extrusion} operator $E$. To achieve this, it suffices  to reshape the channel dimension of the 2D output~($N\times C$) into $N\times C$, giving height dimension to the 2D feature map. In contrast to voxel-based approaches, we compute a 3D feature grid only at the last layer while keeping our intermediate features in 2D, using computationally efficient 2D convolutions only.

We rationalize the generator's design choices in Sec.~\ref{sec:representation}.

\subsection{Rendering and Upsampling}
\label{sec:render}
Given the generated feature volume $G(\textbf{L},\textbf{s}),$ we can query continuous neural field value at any query 3D point $\textbf{p}$ by passing its tri-linear
interpolated feature $\lambda(\textbf{p})$ to a small MLP $\phi(\cdot)$, consisting of a single hidden layer of 64 hidden dimension and softplus activation.
The outputs of the MLP $\phi(\cdot)$ are a scalar density and a 32-dimensional feature, where the first three channels are interpreted as RGB. We do not model view-dependent effects following \cite{chan2022efficient}.
We integrate radiance by volume rendering $\mathcal{R}(.)$ and generate the image
\begin{equation}
\mathcal{I}^\text{low-res}_\gamma = \mathcal{R}(G(\textbf{L},\textbf{s}), \phi(\lambda(\textbf{p})), \gamma)
\end{equation}
from a camera viewpoint $\gamma$. We use 48 points along the ray sampled with stratified sampling and another 48 points obtained with importance sampling \cite{mildenhall2020nerf}.
We set the volume rendering resolution to $64^2$ which provides a reasonable trade-off between computational costs and (post-upsampling) multi-view consistency.

\paragraph{Upsampling}
Volume rendering at our target image resolution of $256^2$ is computationally expensive, so we use the popular 2D super-resolution module (i.e. dual discrimination) of EG3D \cite{chan2022efficient}, which is known to encourage multi-view consistent renderings.
The upsampled image
\begin{equation}
\mathcal{I}_\gamma=\textit{upsample}(\mathcal{I}^\text{low-res}_\gamma,\textbf{s})
\end{equation}
is a function of the volume rendered image $\mathcal{I}^v_\gamma$ and the style code $\textbf{s}$, as we use the StyleGAN2 network for upsampling. 

\begin{figure}
\centering
\centerline{\includegraphics[width=1.0\linewidth]{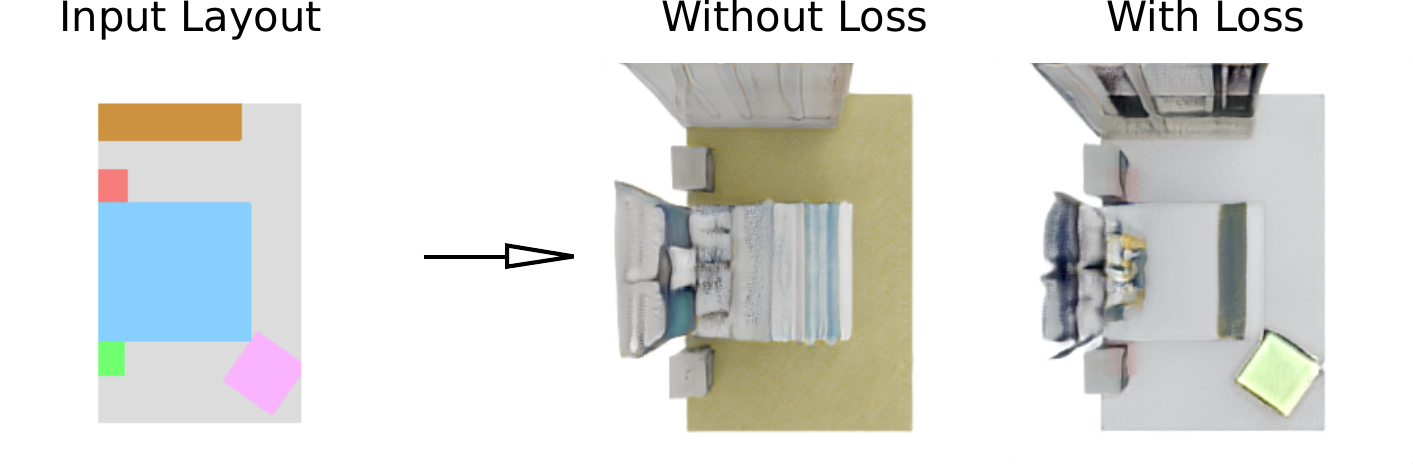}}
\caption{\textbf{Layout consistency loss} -- We often observe objects from the input layout missing in output renderings. We show that using the layout consistency loss helps reduce such cases.}
\label{fig:semantic}
\vspace{-0.3cm}
\end{figure}

\subsection{Discriminator Architecture}\label{sec: discriminator}
Our generator is trained with an adversarial loss that involves co-training a discriminator $D(\cdot)$, which takes real and fake images and predicts their labels. Our discriminator architecture follows that of StyleGAN2 \cite{karras2020analyzing} and takes input as the concatenation of two $256^2$ images following the dual-discrimination scheme \cite{chan2022efficient}.

\begin{figure*}[t]
\begin{center}
\centerline{\includegraphics[width=1.0\linewidth]{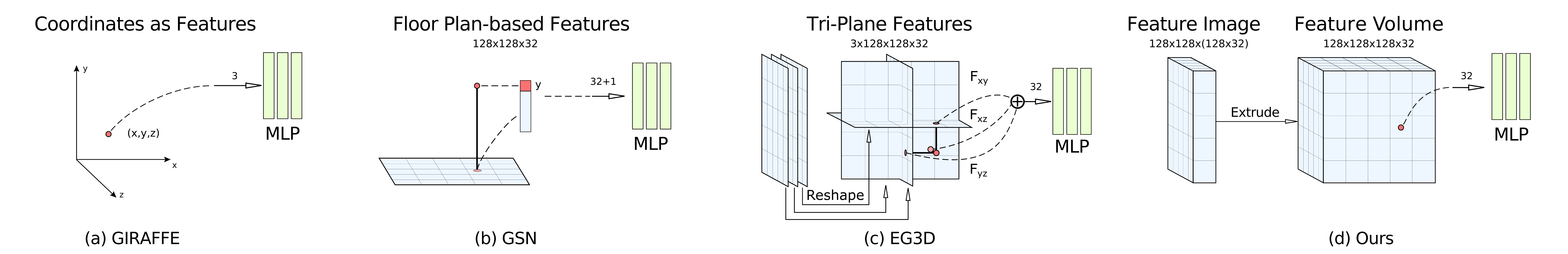}}
\caption{
\textbf{Conceptual Analysis} -- We visually illustrate the difference between the proposed 3D neural field architecture (d), to that used in GIRAFFE~\cite{niemeyer2021giraffe}, GSN~\cite{devries2021unconstrained}, and EG3D~\cite{chan2022efficient}.
Our extrusion operator is computationally much cheaper than those of GIRAFFE or GSN, and compared to EG3D, it better preserves the geometry of the output space: features that are close in $\mathbb{R}^3$ (i.e. volume) have a higher likelihood to be close in $\mathbb{R}^2$ (i.e. image/generator space); given the limited receptive field of 2D convolutions, this results in higher-quality generation.
}
\label{fig:representations}
\end{center}
\vspace{-0.7cm}
\end{figure*}

\paragraph{Enforcing Layout Consistency}
While layout conditioning provides compositional guidance to the generator, we observe that some objects from the input layout are occasionally missing from the final rendering, as shown in Fig.~\ref{fig:semantic}.
To address this, we introduce a semantic layout consistency loss during training that encourages that the generated scene features rendered from the top-down view are consistent with the input layouts.

Specifically, let us define the $xz$-plane as the floorplan and the $y$-axis the up vector. We want to create a 2D image~$\textbf{S}$ on the $xz$-plane that summarizes the generated feature $\mathbf{F}$ from the top-down view. For each pixel in $\textbf{S}$ we sample $k$ number of equidistant points along the height ($y$) axis. Then, we perturb the sampled points with a small Gaussian noise and extract features from those points with tri-linear interpolation. The resulting image $\textbf{S}$ has dimension $N\times N \times (k\times C)$, which is passed to a segmentation U-Net that predicts a semantic label for each pixel. Here, we reuse the discriminator $D(\cdot)$ and attach a decoder network to convert it to a U-Net structure. Besides the adversarial process, the discriminator additionally takes $\textbf{S}$ and outputs semantic segmentation, which is then compared against the input label map $\textbf{L}$ via $\mathcal{L}_{layout}$, a standard cross entropy loss.

\subsection{Training}
We build on top of recent 3D GAN techniques to train our generator by encouraging the neural field renderings from sampled camera viewpoints via adversarial losses.
Specifically, we sample style code $\textbf{s}$ and camera pose $\gamma$ from a prior distribution $p(\gamma)$ and render through the generated neural fields to obtain a fake image $\mathcal{I}_{\gamma}(\textbf{L},\textbf{s}).$
The discriminator takes as input the fake/real images, and outputs the predicted labels.
The two networks are trained via the standard min-max optimization~\cite{goodfellow2014generative}.

\paragraph{Training Objectives}
Our overall training objective comprises the adversarial training loss with R1 regularization loss~\cite{mescheder2018training} and our proposed layout consistency loss of Sec.~\ref{sec: discriminator}, which are weighted equally:
\begin{equation}
\mathcal{L} = \mathcal{L}_{adv} + \mathcal{L}_{R1} + \mathcal{L}_{layout},
\end{equation}
which we minimize by updating the weights of generator~$G(\cdot)$, U-Net backbone~$U(\cdot)$, MLP network~$\phi(\cdot)$, and the extended U-Net discriminator~$D(\cdot)$.


\subsection{Conceptual Analysis}
\label{sec:representation}
As our architecture applies discriminators on the output of the generator, we ought to design a generator architecture that strikes an appropriate balance between \textit{computational requirements} and \textit{3D geometric inductive bias}, both are generally correlated with the visual quality of generated results; see Fig.~\ref{fig:representations} for an overview.

\paragraph{Computational requirements}
While neural implicit representations have greatly advanced 3D generative modeling, classical coordinate-based  implicit representations~(Fig.~\ref{fig:representations}a) require the use of large multi-layer perceptrons~(MLPs).
This incurs in high computational complexity, as every input point evaluates the entire MLP, as well as high memory requirements, as gradients are back-propagated through \textit{all} pixels.

As such, several implicit-explicit hybrid representations have been proposed to pre-load the computational overhead to the generation of explicit features by storing them on regular grids~\cite{fastnerf,kilonerf,nglod}.
Neural field values of query points are obtained by linear feature interpolations, followed by processing with smaller MLPs. Applying these ideas to generative modeling, one can generate 3D features using 3D CNNs as in \cite{xu20223d}; however, 3D convolutions quickly become prohibitively expensive due to the curse of dimensionality.

Recently, GSN~\cite{devries2021unconstrained} suggests adopting planar grids to achieve efficient generation via the use of 2D~\textit{floor-plan} features~(Fig.~\ref{fig:representations}b).
They define the neural field via an MLP that takes the concatenation of the floor-plan projected features and height coordinates.
Since the height-wise information needs to be~``generated'' by the MLP based on the projected 2D features, the heavy lifting is still done by the MLP network, which leads to (prohibitively) large MLP size ~\cite{devries2021unconstrained}.

Conversely, we first generate a 2D feature map, using a 2D U-Net architecture and then {\em extrude} them into 3D volumetric features~(Fig.~\ref{fig:representations}d), thereby pre-computing the height-wise features.
Our 2D-to-3D extrusion strategy enables us to leverage 2D CNNs, and a much smaller MLP to interpret the voxel features.

\paragraph{Geometric inductive bias}
Similar to our approach, tri-plane representations \cite{peng2020convolutional, chan2022efficient} (Fig.~\ref{fig:representations}c) encode 3D information of all axes, allowing a dramatic reduction of the MLP size.
However, these features are jointly generated from a standard 2D CNN and reshaped into three separate planes, leading to the processing of the three planes with very different Euclidean positions via local convolutions.
Moreover, as the scale of the scenes increases, the 2D plane features become less descriptive since completely different objects in the scene share the same plane-projected features.

In contrast, our 2D-to-3D extrusion strategy leverages efficient 2D operations to output 2D feature images, whose individual pixels encode vertical scene information in the height dimension. 
Applying local convolutions on the feature image allows associating geometrically neighboring features, resulting in higher quality results, as empirically validated in our experimental evaluation.

\section{Experiments}

In this section, we provide extensive evaluations of our model on multi-object datasets and compare the results with the most relevant baselines.
Additional results as well as implementation details are provided in the supplementary.

\paragraph{Dataset: 3D-FRONT}
We conduct experiments on \threedfront \cite{fu20213d} bedrooms and living rooms, following the same pre-processing steps as \cite{paschalidou2021atiss}.
To define a camera pose distribution, we consider sampling cameras not inside or too close to the objects to encourage that the majority of the scene's content can be captured. 
To this end, we perform distance transform on the object bounding boxes of the layouts and sample camera locations with sufficiently high distance values with constant heights.
 The orientation is set toward a dominant object in the scene, i.e., beds for bedrooms and the largest object in the scene for living rooms.
We filter out the scenes where we cannot sample at least 40 unique camera poses, resulting in a total of~5515 bedrooms and~2613 living rooms. For each scene we render images using BlenderProc \cite{denninger2020blender} at $256^2$ resolution.
To generate the conditioning inputs, we render a top-down view of each scene with bounding boxes, where each box is colored based on its semantic class and its local coordinates.

\paragraph{Dataset: KITTI}
To demonstrate the generation capabilities of our model in more challenging real-world scenarios, we also evaluate our model on \kitti \cite{liao2022kitti}.
To render our training images, we use the ground-truth camera poses and intrinsic matrices.
Since KITTI scenes are unbounded (\ie there are no specific boundaries), for a single scene we extract several "sub-scenes'' of size $\text{50m} {\times} \text{10m} {\times} \text{50m}$
and use them instead for training. Furthermore, we discard scenes where the car is turning either left of right. This results in 37691 scenes in total.
To render the semantic masks, used to condition the generation, we render top-down views of the scene with bounding boxes, where boxes are colored based on their semantic class.

\begin{figure}[t]
\begin{center}
\centerline{\includegraphics[width=1.0\linewidth]{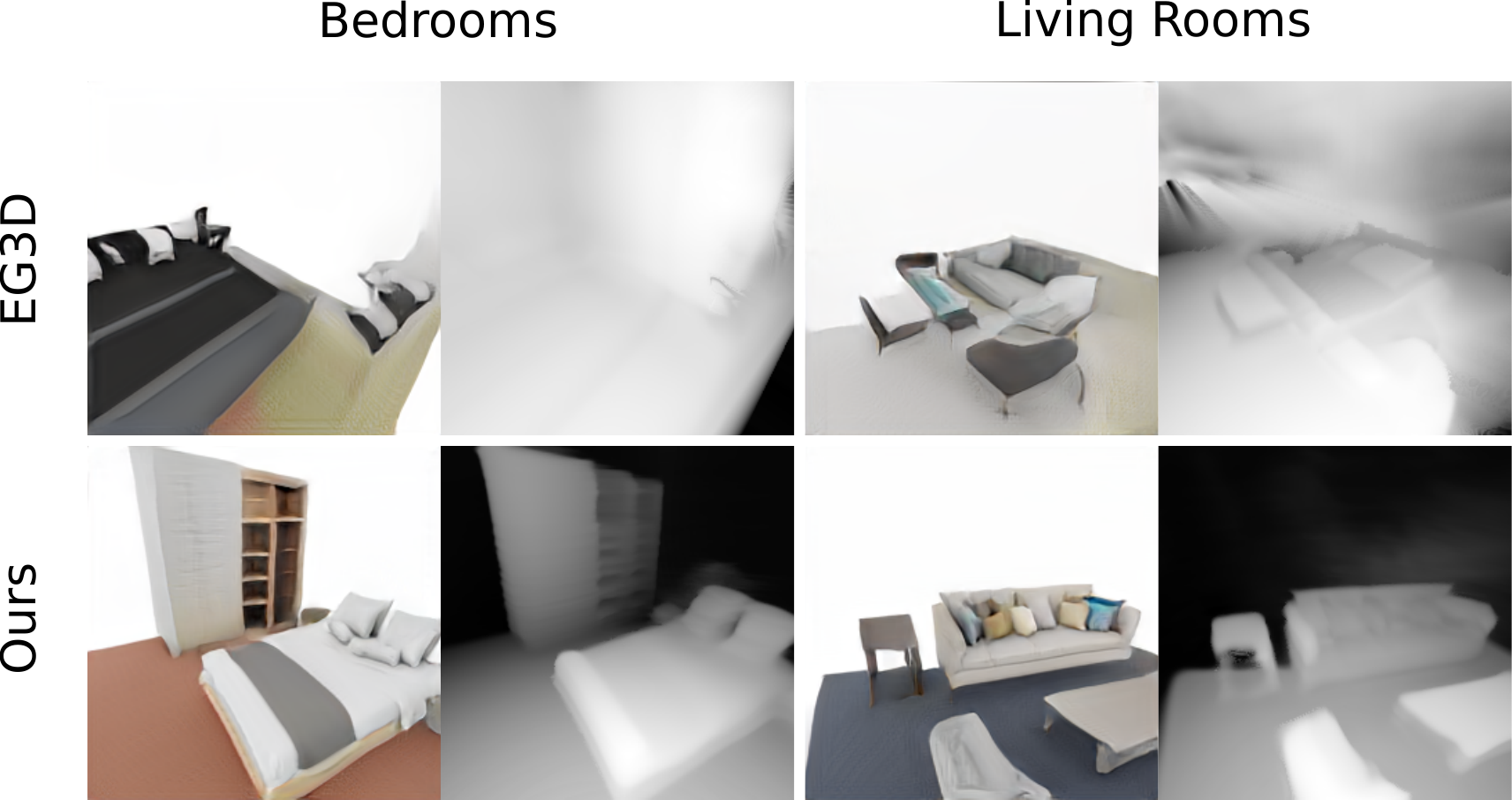}}
\caption{\textbf{Visualization of depth maps} -- extracted from the density fields. Our method is able to generate sharp depth maps in comparison to EG3D, which produces unrecognizable results.}
\label{fig:depth}
\end{center}
\vskip -0.2in
\end{figure}

\begin{figure*}[t]
\begin{center}
\centerline{\includegraphics[width=1.0\linewidth]{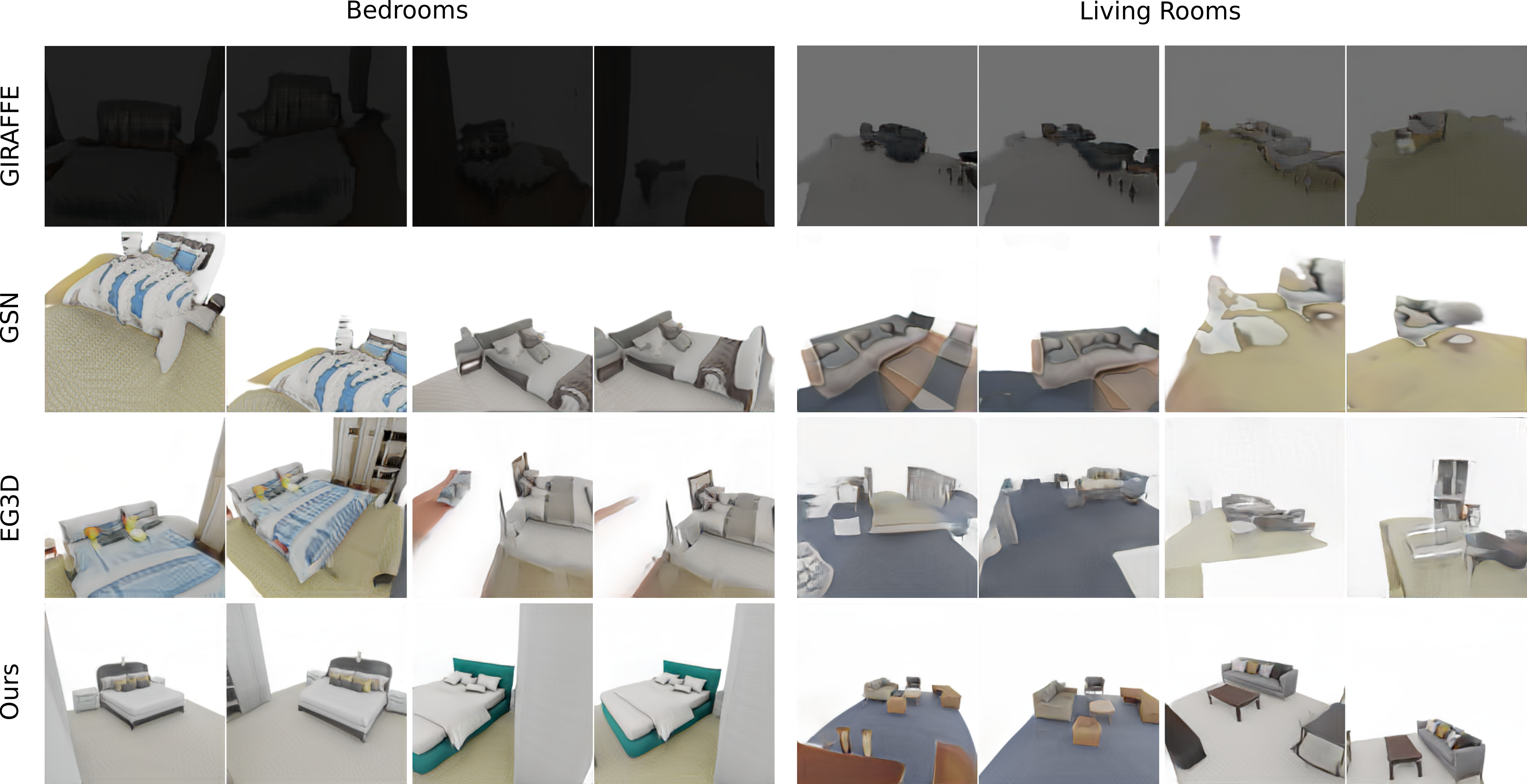}}
\caption{\textbf{{Qualitative comparison on \threedfront bedrooms and living rooms}} -- We show two random viewpoints for each scene. We compare our model with GSN~\cite{devries2021unconstrained}, GIRAFFE~\cite{niemeyer2021campari}, EG3D~\cite{chan2022efficient} and observe that GIRAFFE and GSN fail to output realistic renderings. Although EG3D produces reasonable texture with limited scene compositionality, our method synthesizes more compelling texture and scene structure. Best viewed digitally.}
\label{fig:3dfront}
\end{center}
\vskip -0.2in
\end{figure*}
\begin{table*}[!t]
    \caption{\textbf{Quantitative evaluation} using FID and KID for all methods at $256^2$ pixel resolution on \threedfront bedrooms, \threedfront living rooms, and \kitti.}
    \label{table:main}  
    \vskip 0.15in
    \begin{center}
    \begin{tabularx}{0.785\linewidth}{@{}lcccccccc}
        \toprule
        \multirow{2}{*}{Method} & \multirow{2}{*}{Representation} &\multicolumn{2}{c}{Bedrooms} & \multicolumn{2}{c}{Living Rooms} & \multicolumn{2}{c}{\kitti} \\
        \cmidrule(lr){3-4} \cmidrule(lr){5-6} \cmidrule(lr){7-8}
       & & FID ($\downarrow$) & KID ($\downarrow$) & FID ($\downarrow$) & KID ($\downarrow$) & FID ($\downarrow$) & KID ($\downarrow$) \\
        \midrule
        GIRAFFE & Pure MLP&{141.5} & {127.3}  & {155.7} & {157.5} & {189.0} & {238.3} \\
        GSN & 2D Floor Plan&{73.6} & {43.8}  & {175.4} & {164.9}  & {256.7} & {323.0} \\
        EG3D & Tri-plane&{49.0} & {35.7} & {90.9} & {84.3}  & {78.2} & {82.2} \\
        \midrule
        Ours  & 2D-3D Extrusion&\textbf{28.5} & \textbf{21.3} & \textbf{40.3} & \textbf{34.5} & \textbf{65.6} & \textbf{70.5} \\

        \bottomrule
    \end{tabularx}
    \end{center}
    \vskip -0.1in
\end{table*}
\begin{table}[!t]
    \caption{\textbf{Quantitative ablation studies} on \threedfront bedrooms. We measure the realism of generated 3D scenes without using 2D layout conditioning (i.e., unconditional version of our model) or using the layout consistency loss described in Sec.~\ref{sec: discriminator}. Moreover, we swap out our 3D extrusion representation with the ``floorplan" and tri-plane schemes, proving the advantage of our method.}
    \label{table:ablations}
    \vskip 0.15in
    \begin{center}
    \begin{tabularx}{1.0\linewidth}{@{}X@{}ccc}
        \toprule
        Method & FID ($\downarrow$) & KID ($\downarrow$) \\
        \midrule
        Ours & \textbf{28.5} & \textbf{21.3}\\
        \hspace{1mm} w/o Layout Conditioning & {38.3} & {29.6} \\
        \hspace{1mm} w/o Layout Consistency Loss & {34.2} & {25.3} \\
        \hspace{1mm} w/ GSN's Floorplan Representation & {45.1} & {33.1} \\
        \hspace{1mm} w/ EG3D's Tri-plane Representation & {38.9} & {27.7} \\
        \bottomrule
    \end{tabularx}
    \end{center}
    \vskip -0.1in
\end{table}

\paragraph{Metrics}
We report the Fréchet Inception Distance (FID)~\cite{heusel2017gans} and Kernel Inception Distance (KID)~\cite{binkowski2018demystifying} to measure the realism of the rendered images with respect to the ground truth image distributions. We use 50000 images for \threedfront and the maximum 37691 images for \kitti.

\paragraph{Baselines}
We compare our model with several state-of-the-art methods for 3D-aware image synthesis: GIRAFFE~\cite{niemeyer2021giraffe},  GSN \cite{devries2021unconstrained} and
EG3D~\cite{chan2022efficient}. From our evaluation, we omit GIRAFFE-HD~\cite{xue2022giraffe} as it can only generate single-object scenes and GAUDI~\cite{bautista2022gaudi}, as
the authors have not released any code to train their model.

\paragraph{Quantitative Results}
In Tab.~\ref{table:main}, we provide quantitative evaluations in comparison to the baselines. CC3D demonstrates significant improvements across all metrics and achieves state-of-the-art performance on the scene synthesis task both in indoor and outdoor scenes. 
In comparison to GIRAFFE and GSN, we see that CC3D demonstrates significant improvements with several times smaller FIDs and KIDs, validating that our method better scales to scenes with multiple objects.
Although EG3D shows the most competitive results in comparison to CC3D, our synthesized images are more plausible for both benchmarks.

\begin{figure*}[t]
\begin{center}
\centerline{\includegraphics[width=1.0\linewidth]{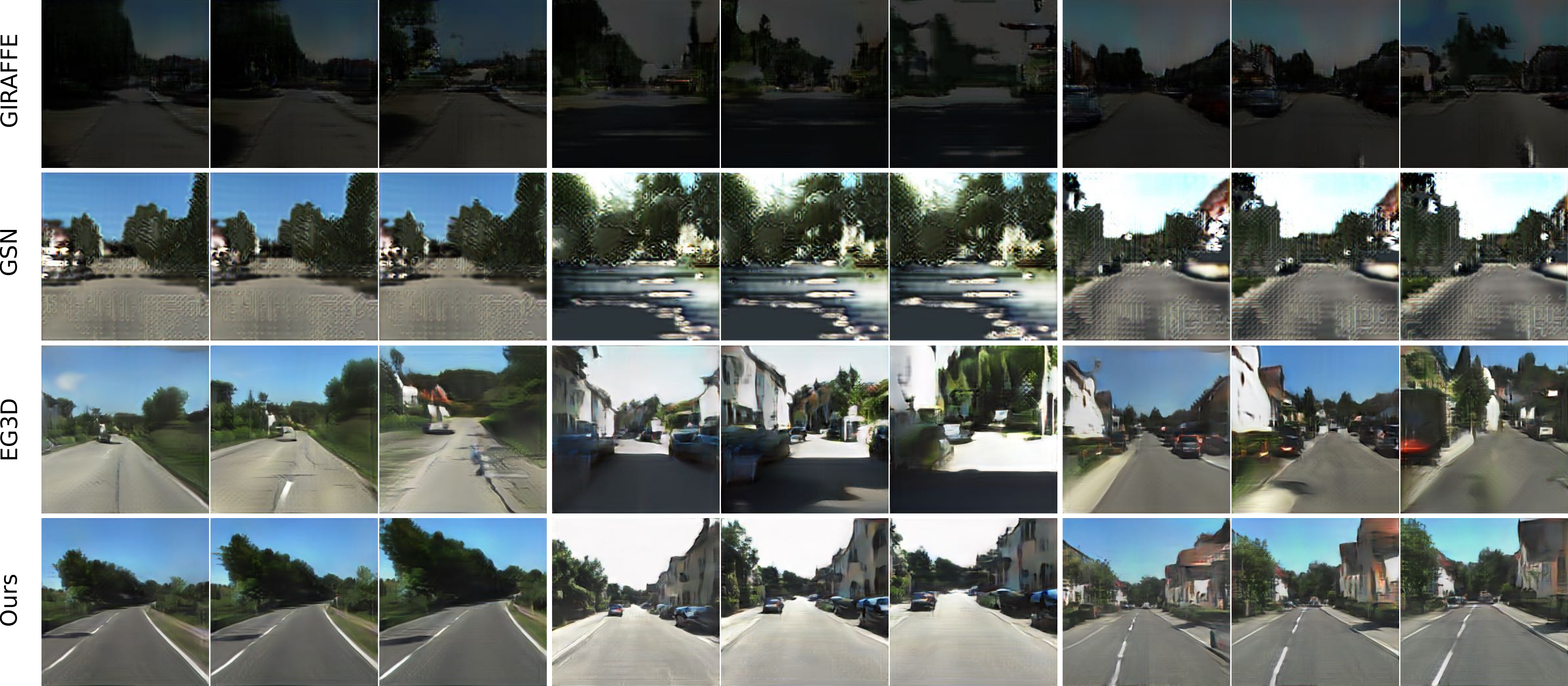}}
\caption{\textbf{Qualitative comparison on \kitti} -- We compare our model with GSN~\cite{devries2021unconstrained}, GIRAFFE~\cite{niemeyer2021campari}, EG3D~\cite{chan2022efficient}.
Although KITTI scenes are more complex, our model can robustly synthesize realistic and diverse outdoor scenes. Best viewed digitally.}
\label{fig:kitti}
\end{center}
\vskip -0.2in
\end{figure*}

\paragraph{Qualitative Results}
In Fig. \ref{fig:3dfront} and Fig. \ref{fig:kitti}, we provide qualitative results for \threedfront and \kitti respectively. As also validated quantitatively in Tab.~\ref{table:main}, our model produces high quality and view-consistent images from different camera poses. In comparison to previous approaches, CC3D synthesizes scene compositions which are more realistic due to our scene layout conditioning. While EG3D shows promising texture quality, the lack of compositionality leads to low-quality underlying scene structures, evidenced by the depth map visualization results (see Fig. \ref{fig:depth}). Notably, all previous methods produce unrealistic scenes for the case of living rooms, unlike CC3D which produces coherent scenes. For the case of \cite{niemeyer2021giraffe}, we observe that it fails to produce plausible scenes \ie most generated scenes are almost completely dark, as also noted in \cite{xu2022discoscene,or2022stylesdf}. Additional results are provided in the supplementary materials.

\subsection{Empirical Analysis}

\paragraph{Layout Conditioning Improves Scene Quality}
Conditioning the generation process with a semantic layout provides compositional guidance to the model. We observe that training an unconditional version of our model leads to a noticeable loss in visual quality as shown in the worse metric scores in Tab.~\ref{table:ablations}. It also aligns with the fact that our conditional method significantly outperforms the existing unconditional GANs, which highlights the importance of providing input conditioning for compositional scene generation.

\paragraph{3D Field Representations}
In Sec.~\ref{sec:representation}, we described how the existing representations for modeling neural fields have trouble modeling large, multi-object scenes. Instead, our 2D-to-3D extrusion method is efficient for using only 2D convolutions and has a strong geometrical inductive bias. 
Hence, to validate our design choice, we substitute 3D field representation with GSN's ``floorplan" and EG3D's tri-plane representations and observe worse performances than ours (Tab.~\ref{table:ablations}), as expected.

\paragraph{Layout Consistency Loss}
As part of our preliminary 2D experiments, we observed that objects are sometimes missing from the output rendering, in particular when there are too many or small objects. Adding our layout consistency loss (Sec.~\ref{sec: discriminator}) during training,  addresses this issue, as shown in Fig. \ref{fig:semantic}. However, we note that the missing object phenomenon still occurs, especially in living room scenes that contain a lot of objects. We will discuss this phenomenon in the supplementary.
Furthermore, we show that using the layout consistency loss improves visual quality (Tab.~\ref{table:ablations}).

\begin{figure}[t]
\begin{center}
\centerline{\includegraphics[width=1.0\linewidth]{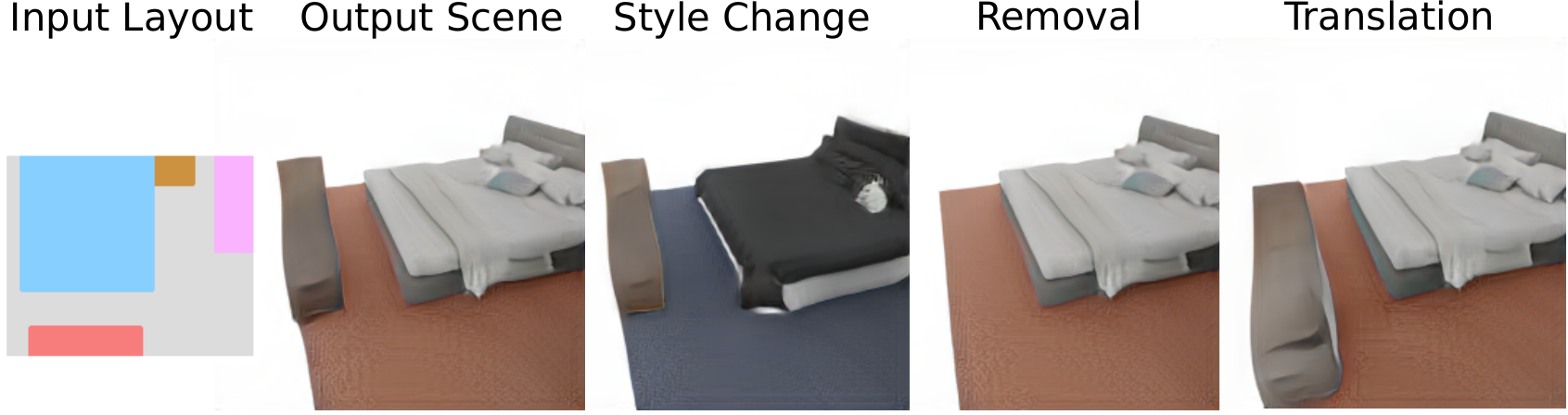}}
\caption{\textbf{Controllable generation of 3D scenes}. Note the change of style, and removal and movement of the tv stand.}
\label{fig:3dfront_3d_edit}
\end{center}
\vspace{-0.5cm}
\end{figure}

\paragraph{Controllable Generations} We showcase that our model enables controlling the 3D scene generation process and supports various editing operations. In Fig.~\ref{fig:3dfront_3d_edit}, we provide examples
of changing the style of the objects, removing objects from the scene and changing the position of an object in the scene.

\section{Discussion}
\label{sec:conclusions}

\paragraph{Limitations}
Even with the semantic consistency loss of Sec.~\ref{sec: discriminator}, there are still missing objects in the generated scenes, especially for large living room scenes with many objects. We believe that tightly enforcing the  generator to closely follow the conditioning is a challenging but important problem that needs to be explored by our community.

Similar to previous 3D-aware GANs, our approach suffers from view-inconsistencies caused by 2D upsampling, which is mostly visible in rendered camera trajectories.
One solution could be to leverage patch-based training to discard the 2D super-resolution module as in~\cite{skorokhodov2022epigraf, son2022singraf}.

The ability to change disentangled latent codes for each object could enable more controlled scene editing, similar to GIRAFFE \cite{niemeyer2021giraffe}.
Furthermore, we observe that the global style code and the input layouts are not completely disentangled, i.e., layout changes often lead to appearance changes. 
Moreover, we rely on a manually-defined camera distribution for each dataset. 
Finally, extending our method to dynamic scenes \cite{bahmani20223d, xu2022pv3d} could enable spatio-temporal control of complex scene generation. We leave addressing the above concerns as future work.

\AT{you could say something about the use of mortom codes to further increase the correlation between CNN feature proximity and geometric proximity? this could be a nice follow up?}

\paragraph{Conclusions}
In this work, we present a conditional 3D GAN, dubbed {\em CC3D}, that can compositionally synthesize complex 3D scenes, supervised only from unstructured image collections and scene layouts. We show that our 2D-conditioned 3D generation technique, along with our novel 3D field representation, enables high-quality generation of multi-object scenes. With CC3D, we can set the layouts of realistic 3D scenes that can be rendered from arbitrary camera trajectories, opening up a research direction towards controllable and scalable 3D generative technologies.

\section{Acknowledgements}

We thank Ahan Shabanov and Yiyi Liao for helping with the 3D-FRONT and KITTI-360 dataset preprocessing. This work was supported by the Natural Sciences and Engineering Research Council of Canada (NSERC) Discovery Grant and Compute Canada / Digital Research Alliance of Canada.
Furthermore, it was supported in part by ARL grant W911NF-21-2-0104, a Vannevar Bush Faculty Fellowship, a
gift from the Adobe Corporation, a PECASE by the ARO, NSF award
1839974, Stanford HAI, and a Samsung GRO.
Despoina Paschalidou was supported by the Swiss National Science Foundation under grant number P500PT\_206946. 

{\small
\bibliographystyle{ieee_fullname}
\bibliography{bibliography_long,bibliography}
}

\clearpage
\appendix
\section*{Appendix}
\section{Video Results}
\label{sec:video_results}

\textbf{We urge readers to view the additional video generation results on the 3D-FRONT and KITTI scenes.} We show our novel rendering videos in comparison to EG3D~\cite{chan2022efficient} and GSN~\cite{devries2021unconstrained}.
The videos are best viewed by opening our project page.
We note that our model consistently produces more realistic scenes that are view-consistent, whereas the generated scenes from our baselines have more artifacts. Furthermore,
to showcase that our generated scenes have more accurate geometries, we visualize the corresponding depth maps for our generated scenes. We observe that in comparison to EG3D~\cite{chan2022efficient} our depth maps are consistently plausible, less noisy and better capture the 3D object geometries as well as the backgrounds.

More specifically, we note that the results of EG3D on the bedroom datasets suffer not only from the lower visual qualities but also from the structural errors of creating two symmetric scenes about the feature plan. Moreover, in the more challenging living room scenes with many objects, GSN and EG3D fail to generate recognizable objects. In contrast, CC3D consistently synthesizes high-quality, view-consistent scenes. Finally, on the \kitti scenes, EG3D and GSN suffer from significant artifacts in both RGB and depths with large view changes.
\section{Layout Consistency}
\label{sec:app_consistency}

We conduct an experiment to measure the layout consistency between the input semantic layouts and output top-down renderings of the scenes. To this end, we render all 5515 bedroom scenes with our trained models from top-down views. To test the effects of the layout consistency loss, as introduced in Sec. 3.3 of the main document, we obtain two generated top-down images, one rendered from the model trained with and the other one trained without the layout consistency loss. We pair the top-down renderings with ground truth semantic labels and separate the obtained datasets into train and test splits using a ratio of 0.85. Then, we train a DeepLabv3 \cite{chen2017rethinking} semantic segmentation model to predict the ground truth semantic segmentations from the rendered town-down images. Essentially, the more consistent the layouts and the top-down renderings the easier it becomes for the segmentation model to learn. Therefore, better test performance likely indicates higher consistencies. In Tab. \ref{table:semantic}, we indeed observe an improved IoU numbers when training our model with layout consistency loss.

In addition, we conducted another experiment where we trained Faster R-CNN \cite{ren2015faster} on top-down renderings to detect instances w.r.t. the input layout and evaluate using average precision (AP) and average recall (AR) with the COCO evaluator. The consistency metrics are generally high and we observe improved numbers (AP: 0.774, AR: 0.809) against the model without the consistency loss (AP: 0.719, AR: 0.747) in Eq. 4 of the main paper.

\begin{table}[!htbp]
    \caption{\textbf{Quantitative measures of semantic consistency} -- between input layouts and top-down rendering outputs on \threedfront Bedroom scenes.}
    \label{table:semantic}
    \vskip 0.15in
    \begin{center}
    \begin{tabularx}{1.0\linewidth}{@{}X@{}cccc}
        \toprule
        Method & IoU $\uparrow$ \\
        \midrule
        Ours & 0.7182\\
        \hspace{3mm} w/o Layout Consistency loss & 0.6593\\
        \bottomrule
    \end{tabularx}
    \end{center}
    \vskip -0.1in
\end{table}
\section{Implementation Details}
\label{sec:impl_details}
Below we share the implementation details of our training and testing pipelines. To promote reproducibility, we will release the code to the public upon acceptance.

\paragraph{Baselines}
\label{sec:baselines}
 We use the R1 regularization \cite{mescheder2018training} across all datasets and models. We replace the sphere-based camera sampling in EG3D~\cite{chan2022efficient} and GIRAFFE~\cite{niemeyer2021giraffe} to allow freely moving cameras, which are significantly challenging to learn from. As stated in the main text, we sampled the cameras by first performing the distance transform and then randomly sampling locations where the distance value is above the threshold. For bedrooms, we orient the camera toward a randomly sampled point within the bed bounding box and for living rooms, we used the largest object in the scene. For GSN~\cite{devries2021unconstrained}, we do not use depth maps as supervision and train on single-view image collections in the same setup as all other models. We train all models using 3,000,000 images for all datasets and models with batch size 32.

From our qualitative evaluations in the main paper, we note that the generated scenes of GIRAFFE \cite{niemeyer2021giraffe} are quite dark.
 We observed that the low brightness of GIRAFFE results can be prevented with higher R1 regularization, however, this led to many samples being completely black and higher FID scores. Hence, we used the regularization factor which obtains the lowest FID score for a fair comparison.
 
 Even though we show several generated scenes using GSN~\cite{devries2021unconstrained}, we would like to note that training GSN on \threedfront living rooms and \kitti, even with hyperparameter tuning was not trivial and resulted in the model collapsing. We hypothesize that GSN requires depth supervision for better performance on these datasets.

 \paragraph{Training Details}
\label{sec:train_details}
We build our model on top of the EG3D~\cite{chan2022efficient} pipeline, hence
we follow their implementation protocol and hyperparameters unless otherwise stated. We
use a discriminator learning rate of 0.002 and a generator learning rate of 0.0025 with Adam optimizer using $\beta_1 = 0$, $\beta_2 = 0.99$, and $\epsilon = 10^{-8}$. Our mapping network transforms a 512 latent code vector into an intermediate latent code with 2 fully connected layers of dimension 512. We do not apply any pose conditioning to the generator or the discriminator networks. As stated above, we abandon the sphere-sampling of camera locations. Our model takes around 2 days to converge using 4 NVIDIA V100 with 32 GB memory. 

 \paragraph{Semantic Layout Details}
\label{sec:layout}
As mentioned in the main document, we process datasets-dependent layouts $\textbf{L}$ as conditional inputs to our model.
To prepare the input $\textbf{L}$ we discretize the provided semantic 2D floor plans onto the 2D grids. 
For indoor scenes, i.e., 3D-FRONT bedroom and living room scenes, we simply project the 3D bounding boxes of the scenes onto the ground plane and encode the semantic class of each pixel using a one-hot vector and a binary room layout mask. The semantic feature channels are concatenated with the local coordinates of each bounding box~(origin at the left-top corner of the bounding boxes at their canonical orientations), providing orientation information to the subsequent U-Net. In total we concatenate features comprising i) a binary mask of the room layout (1), ii) local coordinates of each object (3), iii) one-hot embedding of the semantic layout (16), and iv) the global latent noise (512). This results in a feature grid with 532 channels for \threedfront. We directly obtain a semantic floorplan representation for outdoor scenes by rendering the 3D semantic annotations from a top-down view. Hence, the semantic maps are pixel-based as opposed to bounding-box-based. For \kitti we concatenate features comprising i) the top-down rendered semantic layouts encoded as a one-hot feature grid (59) and ii) latent noise (512). This results in a feature grid with 571 channels for \kitti.

\subsection{Architecture Details}

Our U-Net is composed of an encoder and a decoder network, each of which is composed of the building blocks of StyleGAN2.
For the encoder, we use the StyleGAN2 synthesis layers except that we replace  the upsampling with the max-pool downsampling operation and use style-modulation with {\em constant} style code 
(i.e., the encoder network is independent of the sampled style code $\bm{s}$). The downsampling layers are repeated to make the feature resolution $4^2$. We turned off the style modulation with the global latents in order to keep the encoder deterministic and only modulate the decoder part of the U-Net. Our U-Net decoder closely follows the StyleGAN2 architecture and starts with a learnable feature with $4^2$ resolution. We use skip connections to concatenate the encoder features to the intermediate features of the corresponding decoding layer. In contrast to the encoder, we modulate the decoder via FiLM with the per-layer style code that is obtained by processing the global style noise vector $\textbf{s}$ with the StyleGAN2 mapping network.

The discriminator architecture follows that of StyleGAN2, except that we attach a decoder network that is symmetric with the encoder network. The two network components are connected via skip connections as in a typical U-Net. We use $k=8$ for the segmentation branch.

\begin{table}[!t]
    \caption{\textbf{Quantitative ablation studies} on \threedfront living rooms. We measure the realism of generated 3D scenes without using 2D layout conditioning (i.e., unconditional version of our model) or using the layout consistency loss described in Sec.~\ref{sec: discriminator}. Moreover, we swap out our 3D extrusion representation with the ``floorplan" and tri-plane schemes, proving the advantage of our method.}
    \label{table:ablations_living}
    \vskip 0.15in
    \begin{center}
    \begin{tabularx}{1.0\linewidth}{@{}X@{}ccc}
        \toprule
        Method & FID ($\downarrow$) & KID ($\downarrow$) \\
        \midrule
        Ours & \textbf{40.3} & \textbf{34.5}\\
        \hspace{1mm} w/o Layout Conditioning & {60.1} & {54.1} \\
        \hspace{1mm} w/o Layout Consistency Loss & {44.7} & {38.0} \\
        \hspace{1mm} w/ GSN's Floorplan Representation & {65.6} & {59.0} \\
        \hspace{1mm} w/ EG3D's Tri-plane Representation & {69.3} & {60.8} \\
        \bottomrule
    \end{tabularx}
    \end{center}
    \vskip -0.1in
\end{table}

\section{Discussions}
\subsection{Note on Tri-Plane Results}
\begin{figure*}[t]
\begin{center}
\centerline{\includegraphics[width=0.5\linewidth]{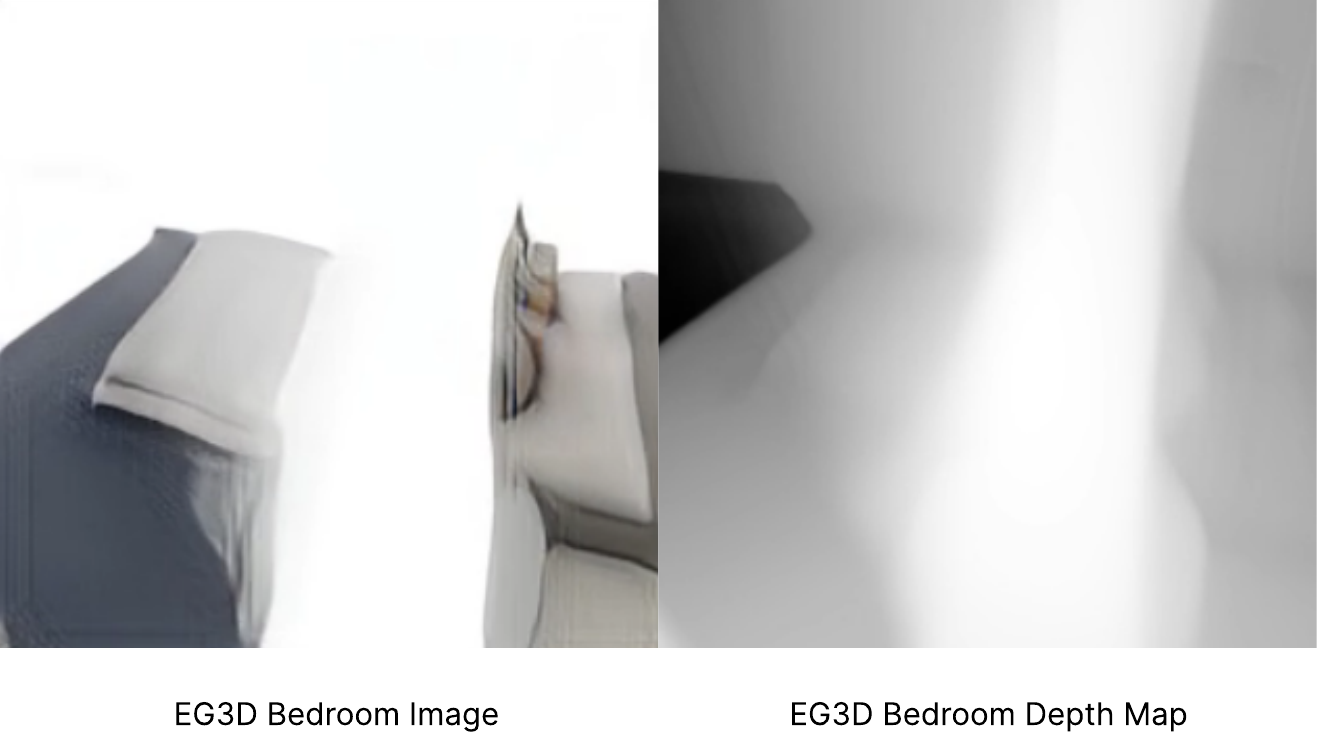}}
\caption{\textbf{Failure case of EG3D \cite{chan2022efficient}} -- on 3D-FRONT bedroom. We notice that the tri-plane representation induces replicating the scenes that are symmetric about one of the feature planes.}
\label{fig:eg3d_failure}
\end{center}
\vskip -0.2in
\end{figure*}

As discussed in the main text, we hypothesize that the tri-plane representation is conceptually not ideal for representing large-scale scenes due to weak geometric inductive bias when generating the tri-plane jointly. Moreover, as the scene gets larger, the same plane-projected features are used to describe totally different objects in a scene, which hampers the representational power of tri-planes. Indeed, in the included video websites, one can observe that the EG3D results contain artifacts where the scene contains two bedrooms, symmetric about one of the three planes (see Fig.~\ref{fig:eg3d_failure}). 
\quad
We also note that the ablated version of our conditional model using the Tri-plane representation suffers from severe layout inconsistencies.
That is, we observe that the input layout is almost completely ignored and the output scenes have almost no resemblance to the input layouts, which clearly indicates that the tri-plane representation lacks geometric inductive bias in our use-case.

\subsection{Note on ``Floorplan" Results}
We discussed in the main text how the ``floorplan'' representation requires a larger MLP because the vertical information needs to be decoded, or ``generated'' by the MLP network. Indeed, we observe that GSN \cite{devries2021unconstrained} adopts an 11-layer MLP with 128 channels in the hidden layers. In comparison, EG3D and our representation require using a two-layer MLP with 64 channels in the hidden layer. Approximately, our MLP network size (in number of weights) is less than 20 times smaller than that of GSN's. 

In our ablation study in the main text (Tab. 2), we swap our ``extrusion'' representation with a ``floorplan'' representation. Here, to make the comparison fair, we used the same two-layer MLPs for the experiment. Note that, while increasing the size of the MLP might improve its performance, it comes at a significant cost of computational resources.

\section{Additional Results}

\begin{figure*}[t]
\begin{center}
\centerline{\includegraphics[width=1.0\linewidth]{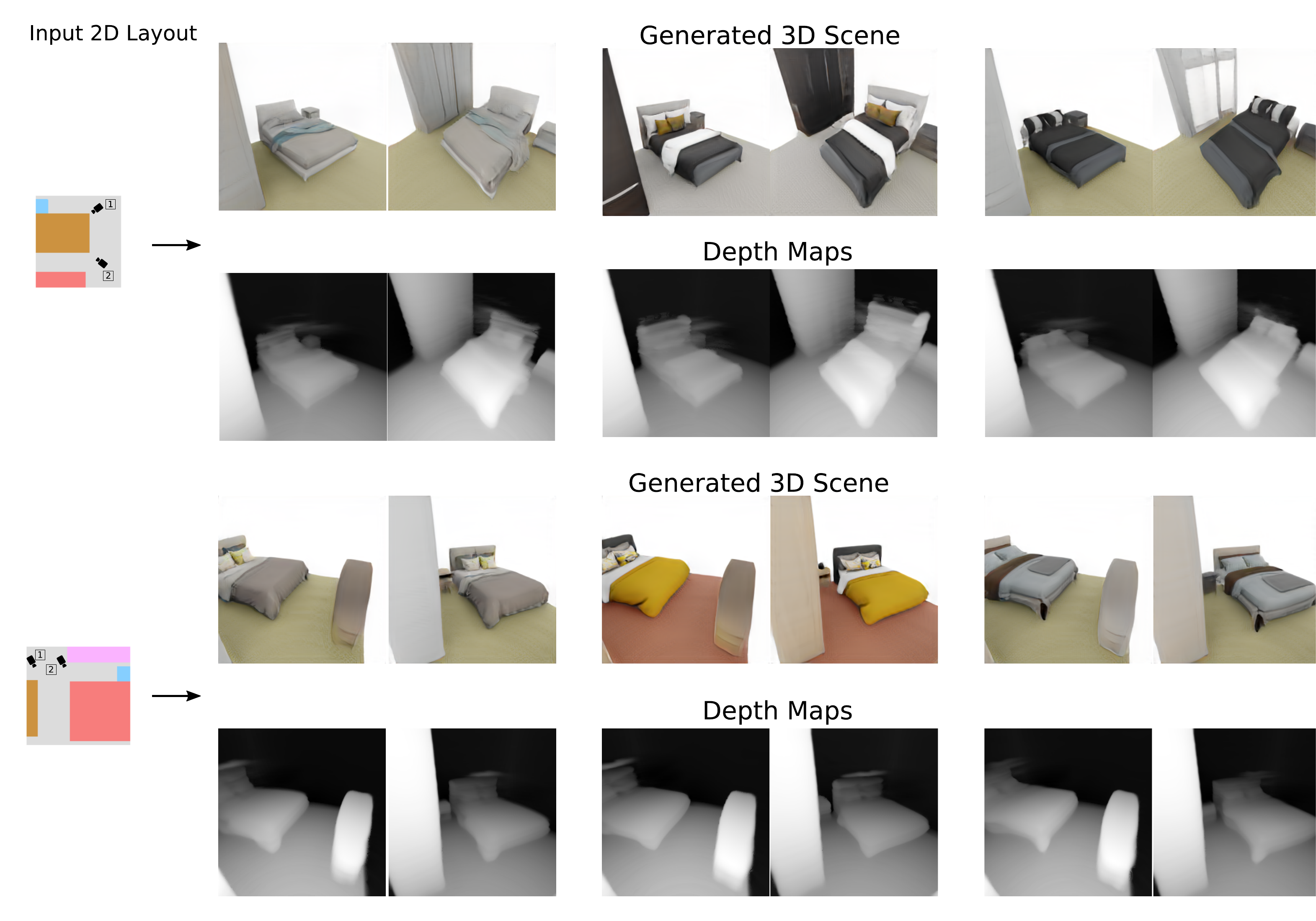}}
\caption{\textbf{Additional results on 3D-FRONT.} We visualize the conditional generation results on 3D-FRONT bedrooms with varying latent codes (shown in three styles). Note that changing the global latent codes results in a change of general styles.}
\label{fig:additional_beds}
\end{center}
\vskip -0.2in
\end{figure*}

\begin{figure*}[t]
\begin{center}
\centerline{\includegraphics[width=1.0\linewidth]{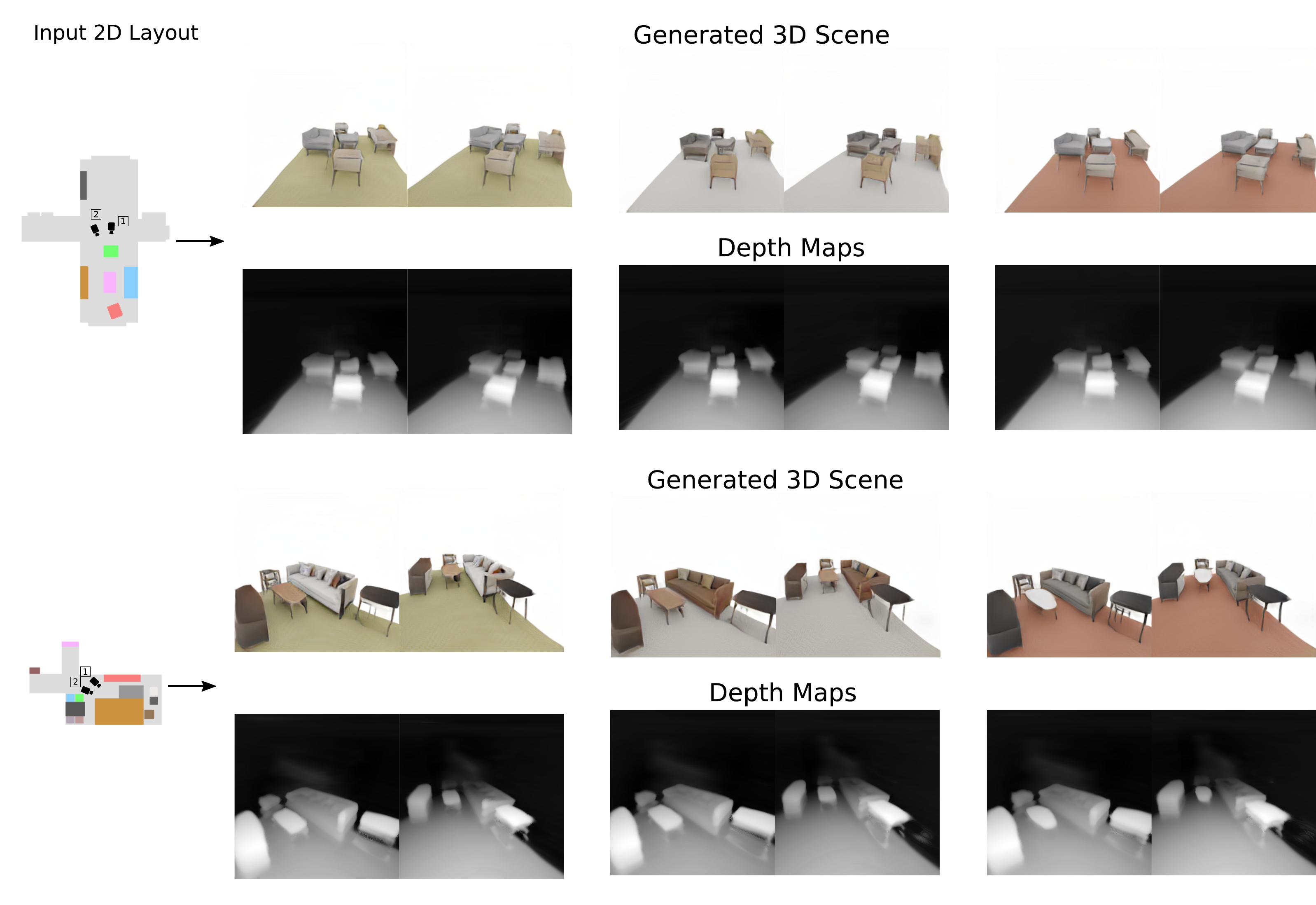}}
\caption{\textbf{Additional results on 3D-FRONT.} We visualize the conditional generation results on 3D-FRONT living rooms with varying latent codes (shown in three styles). Note that changing the global latent codes results in a change of general styles. We notice that for living room scenes the layout conditionings are not perfectly respected. For example, the big sofa bounding box of the second example is splitted into a sofa and a side table.}
\label{fig:additional_living}
\end{center}
\vskip -0.2in
\end{figure*}

\begin{figure*}[t]
\begin{center}
\centerline{\includegraphics[width=0.8\linewidth]{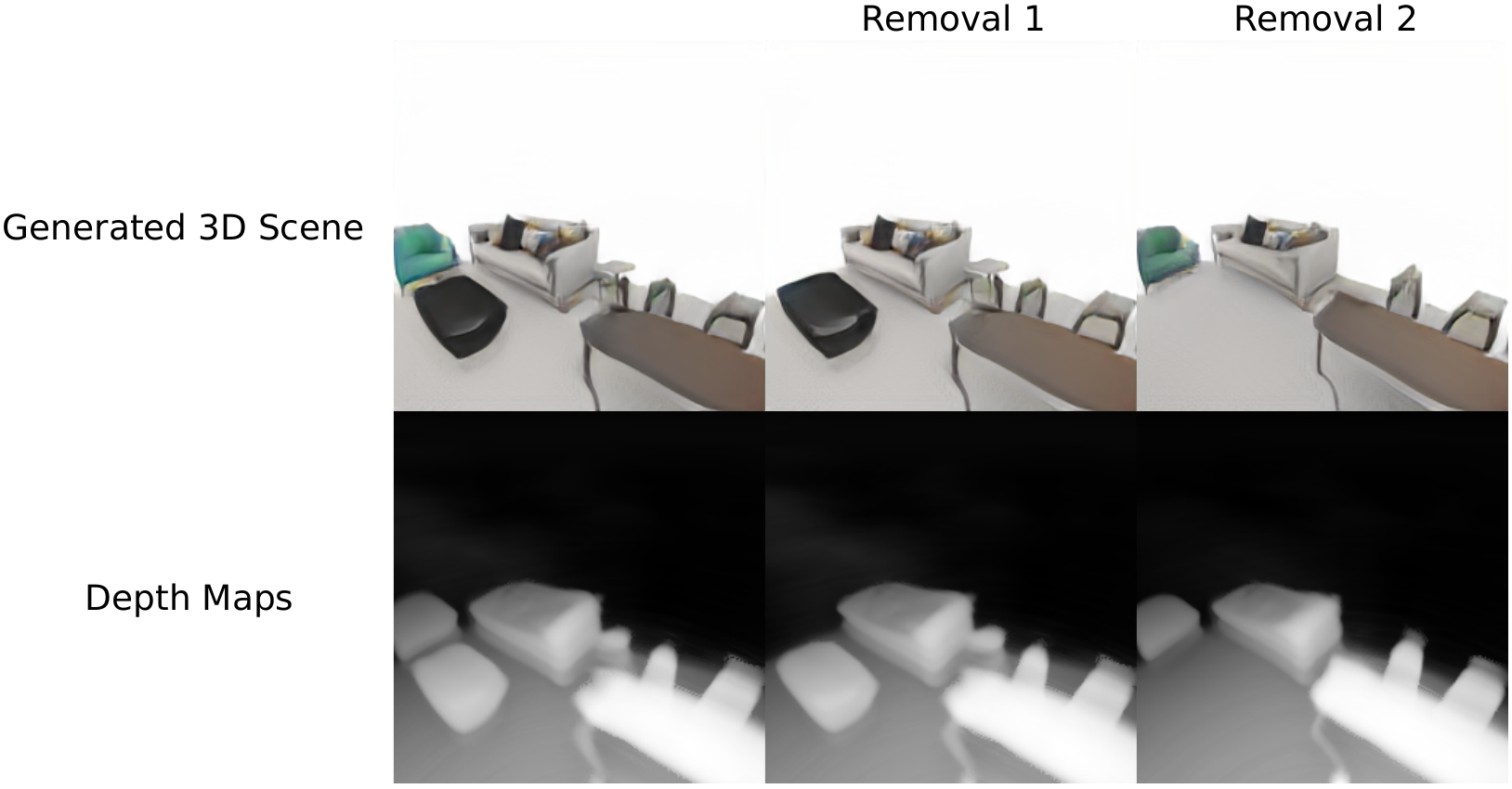}}
\caption{\textbf{Object removal experiment.} We showcase the object removal capability of our approach. Note that from the image on the leftmost column, we can remove the green sofa chair (middle column) and the black coffee table (right column).}
\label{fig:living_removal}
\end{center}
\vskip -0.2in
\end{figure*}

\begin{figure*}[t]
\begin{center}
\centerline{\includegraphics[width=1.0\linewidth]{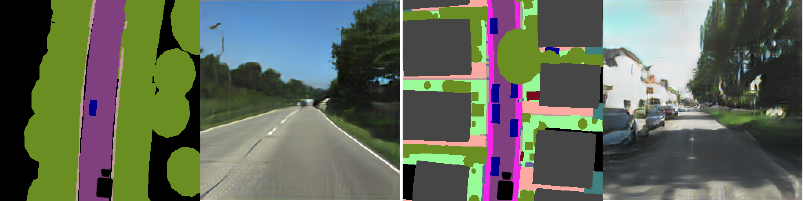}}
   \caption{\textbf{KITTI-360 conditioning.} Layout inputs and generated 3D scenes for KITTI-360.}
\label{fig:kitti_layout}
\end{center}
\vskip -0.2in
\end{figure*}

In Fig.~\ref{fig:additional_beds} and Fig.~\ref{fig:additional_living}, we show additional visualizations of our conditional generation results. Note that the generated 3D scenes generally follow the input layouts. Moreover, we sample three different global latent vectors which, when applied to the generation process, synthesize scenes with different styles. In Fig.~\ref{fig:living_removal}, we demonstrate the object removal capability. Note how we can remove individual objects such as a chair and a coffee table. In Fig.~\ref{fig:kitti_layout} we visualize KITTI-360 layouts and renderings. While our model generates better image quality and view consistency than previous works, we acknowledge that the current model has difficulties closely following complex layouts.

\begin{figure*}[t]
\begin{center}
\centerline{\includegraphics[width=0.8\linewidth]{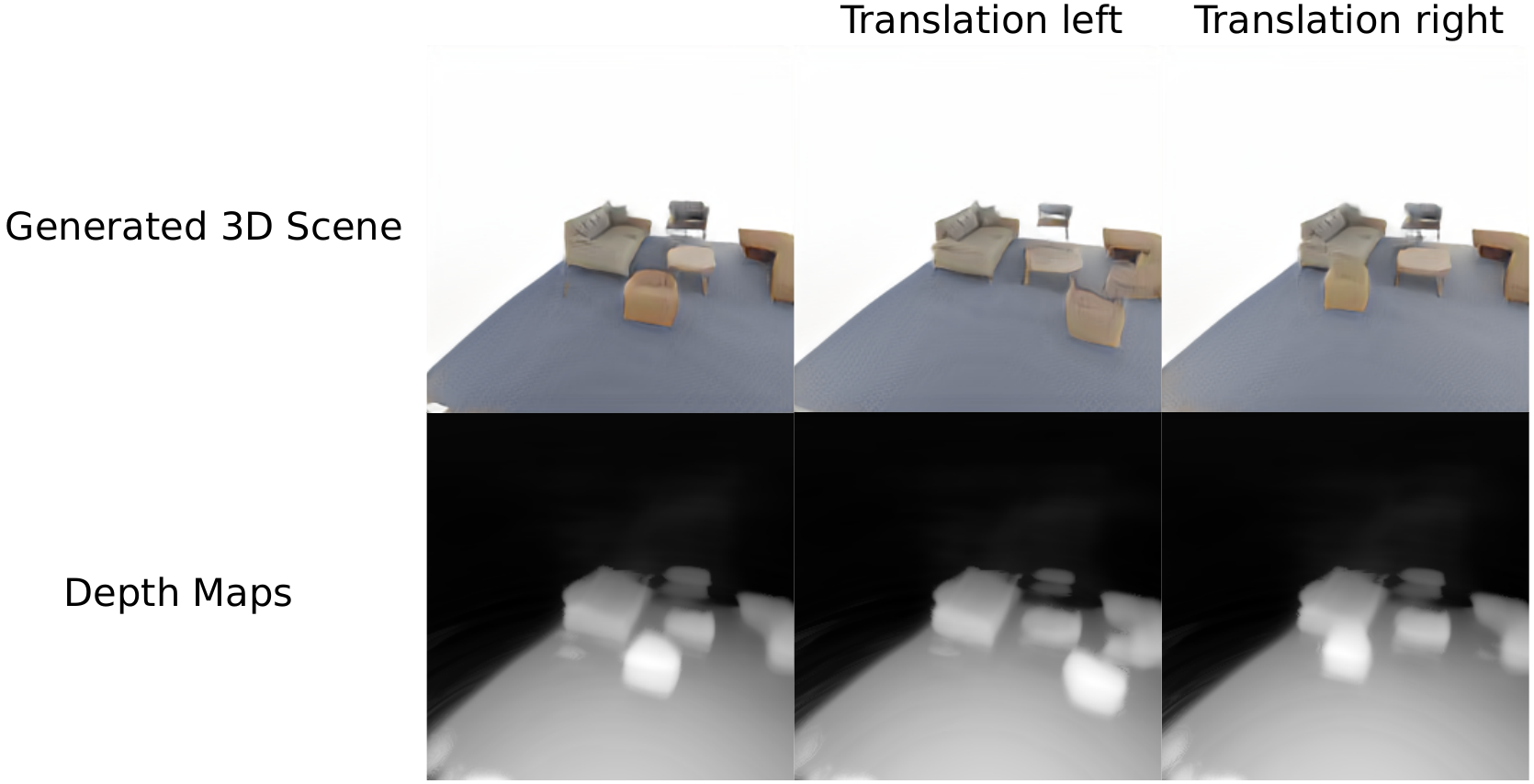}}
\caption{\textbf{Translation experiment.} We showcase the object translation capability of our approach. Note that from the image on the leftmost column, we can translate the object to the left (middle column) and right (right column).}
\label{fig:living_trans}
\end{center}
\vskip -0.2in
\end{figure*}

\end{document}